\documentclass{article}
\usepackage{spconf,amsmath,graphicx}
\usepackage[colorlinks,linkcolor=black,anchorcolor=black,citecolor=black,urlcolor=blue]{hyperref}

\title{Exploiting Modality-Specific Features For Multi-Modal Manipulation Detection And Grounding}
\name{Jiazhen Wang, Bin Liu\sthanks{Corresponding author: \href{binliu:flowice@ustc.edu.cn}{flowice@ustc.edu.cn}}, Changtao Miao, Zhiwei Zhao, Wanyi Zhuang, Qi Chu, Nenghai Yu}
\address{School of Cyber Science and Technology, University of Science and Technology of China}
\begin{document}
%\ninept
%
\maketitle
\begin{abstract}

AI-synthesized text and images have gained significant attention, particularly due to the widespread dissemination of multi-modal manipulations on the internet, which has resulted in numerous negative impacts on society. Existing methods for multi-modal manipulation detection and grounding primarily focus on fusing vision-language features to make predictions, while overlooking the importance of modality-specific features, leading to sub-optimal results. In this paper, we construct a simple and novel transformer-based framework for multi-modal manipulation detection and grounding tasks. Our framework simultaneously explores modality-specific features while preserving the capability for multi-modal alignment. To achieve this, we introduce visual/language pre-trained encoders and dual-branch cross-attention (DCA) to extract and fuse modality-unique features. Furthermore, we design decoupled fine-grained classifiers (DFC) to enhance modality-specific feature mining and mitigate modality competition. Moreover, we propose an implicit manipulation query (IMQ) that adaptively aggregates global contextual cues within each modality using learnable queries, thereby improving the discovery of forged details. Extensive experiments on the $\rm DGM^4$ dataset demonstrate the superior performance of our proposed model compared to state-of-the-art approaches.

\end{abstract}
\begin{keywords}
multi-modal, media manipulation, transformer, modality-specific
\end{keywords}
\section{Introduction}
\label{sec:intro}

\begin{figure*}[htb]
\begin{center}
\includegraphics[width=0.95\textwidth]{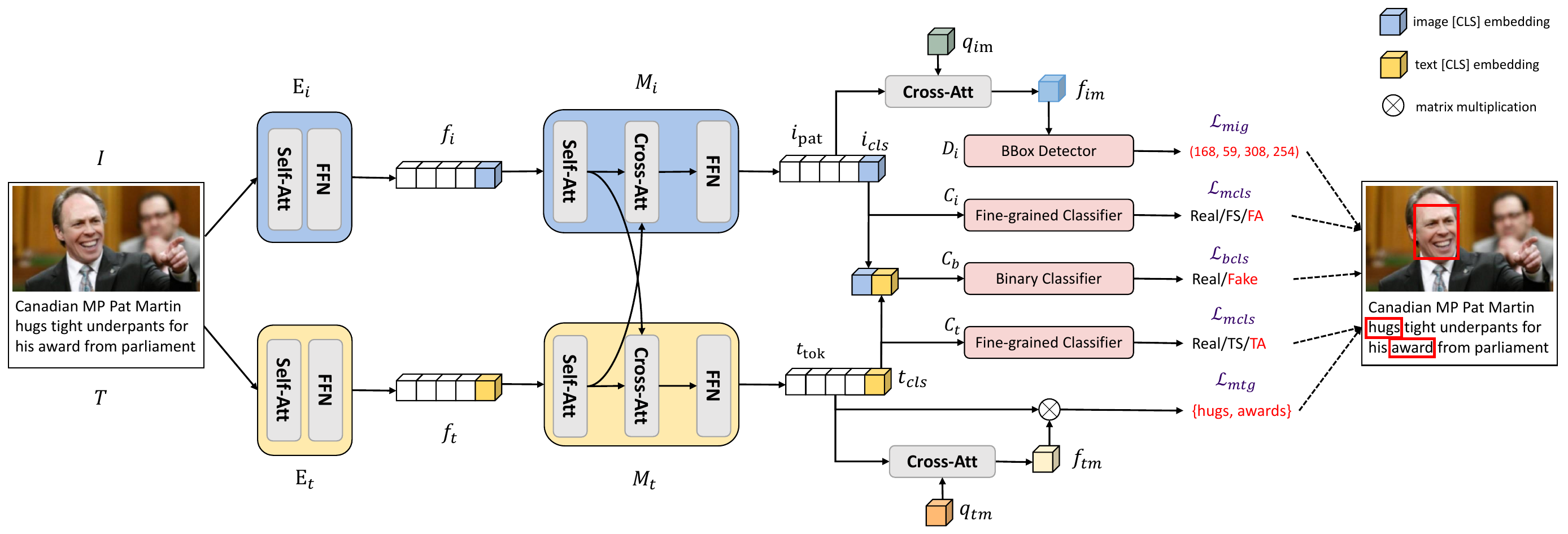}
\end{center}
\caption{The overall architecture of our framework. 1) Image and text features are extracted and fused through uni-model encoders $E_i$, $E_t$, and modality interaction module $M_i$, $M_t$. 2) Decoupled fine-grained classifier $C_i$, $C_t$ and binary classifier $C_b$ take image embedding $i_{cls}$, text embedding $t_{cls}$, and concatenated embeddings $\{i_{cls}, t_{cls}\}$ as inputs, respectively. 3) Image embeddings $i_{pat}$ and text embeddings $t_{tok}$ are separately fed into the implicit manipulation query module and grounding heads.}
\label{Fig.network}
\end{figure*}

The rapid development of deep generative models and large language models has facilitated the easy generation of massive amounts of fake facial images, videos, and synthetic text. These deepfake products \cite{dolhansky2020deepfake, zellers2019defending} have the potential to spread widely on social media. Consequently, this threat has garnered significant attention in the fields of computer vision and natural language processing, leading to the proposal of various methods for detecting fake faces and AI-generated text \cite{miao2021towards, miao2021learning, zhuang2022uia, miao2022hierarchical, miao2023f, zhuang2022towards, tan2022transformer, gehrmann2019gltr}. However, these methods often focus on a single modality. Yet, in everyday life, multi-modal media content in the form of image-text pairs is more prevalent. As a result, multi-modal fake content is more likely to spread widely and cause social problems.

Previous research on multi-modal fake news primarily concentrates on the binary classification problem of determining news authenticity. CMC \cite{wei2022cross} leverages knowledge distillation to capture cross-modal feature correlations during training. CMAC \cite{zou2023cross} combines adversarial learning and contrastive learning to obtain multi-modal fused representations with modality in-variance and clear class distributions. However, these methods \cite{wei2022cross, zou2023cross, ying2023bootstrapping} are unable to determine specific types of manipulation or localize manipulation positions, lacking applicability and interpretability. HAMMER \cite{shao2023detecting} constructs the first dataset for the detection and grounding of multi-modal manipulated content. Furthermore, it proposes a contrastive learning-based approach for modality alignment as well as shallow and deep manipulation reasoning. However, this approach overlooks the importance of modality-specific features, potentially leading to underutilization of the rich information present in each modality and resulting in sub-optimal performance.

In order to exploit modality-specific features, we construct a simple and novel transformer-based framework for multi-modal manipulation detection and grounding. We utilize visual/language pre-trained uni-modal encoders to extract modality-unique features. These features are then fused by dual-branch cross-attention (DCA) to summarize multi-modal information while preserving the individual characteristics of each modality. Furthermore, we introduce decoupled fine-grained classifiers (DFC) to mitigate modality competition, in which one modality is learned well while the other is not fully explored. Additionally, we propose an implicit manipulation query (IMQ) that facilitates reasoning between learnable queries and the global context of images or text, enabling the model to discover potential forged clues.

The main contributions of our paper are as follows: (1) We construct a simple and novel transformer-based framework for multi-modal manipulation detection and grounding. (2) We introduce DCA to fuse modality-unique features while maintaining uni-modal characteristics. Additionally, we design DFC and IMQ to promote comprehensive exploration of each modality and enhance intra-modal interactions, respectively. (3) We conduct experiments on the $\rm DGM^4$ dataset, and the results demonstrate the superiority and effectiveness of our approach.

\section{METHODOLOGY}
\label{sec:methodology}

\subsection{Feature Extraction}
\label{ssec:FE}

In order to capture modality-specific features for images and text, we employ two visual/language pre-trained uni-modal feature extractors, denoted as $E_i$ for images and $E_t$ for text. As illustrated in Figure 1, given an image-text pair $(I, T)$, we divide the input image $I$ into $N$ patches and insert a \texttt{[CLS]} token. Similarly, we segment the input text $T$ into $M$ tokens and insert a \texttt{[CLS]} token. The image features $f_i$ and text features $f_t$ are separately extracted by $E_i$ and $E_t$.

\begin{equation}
f_i = E_i(I), \quad f_t = E_t(T).
\end{equation}

In a single modality, manipulation cues are often subtle and not easily discernible. However, in multi-modal manipulation content, there can be distinctive information that differs between the modalities. To effectively reason about the correlations between images and text, a deep level of interaction and fusion between their features is crucial. We employ a dual-branch cross-attention (DCA) mechanism, as depicted in Figure 1, to guide the interaction between image features $f_i$ and text features $f_t$. Unlike HAMMER \cite{shao2023detecting} which uses single-stream interaction with text features as queries but image features remain unchanged, DCA allows each embedding to capture modality-contextual information while enabling deep interactions with information from the other modality.

The attention function is applied to query ($Q$), key ($K$), and value ($V$) features, each with a hidden size of $D$, as follows:

\begin{equation}
\text{Attention}(Q, K, V) =\text{Softmax}(Q K^T / \sqrt{D})V.
\end{equation}

In DCA, given queries from one modality (e.g., image), keys and values can be taken only from the other modality (e.g., text). We denote two modality interaction modules as $M_i$ and $M_t$, where cross-attention in $M_i$ using image features as queries and cross-attention in $M_t$ using text features as queries. Multi-layer self-attention and DCA combination to achieve complex inter-modal fusion. This can be expressed as

\begin{equation}
M_i(f_i, f_t)=\{i_{cls}, i_{pat}\}, \quad M_t(f_t, f_i)=\{t_{cls}, t_{tok}\}.
\end{equation}

Here, $i_{cls}$ represents the embedding of the image \texttt{[CLS]} token, $i_{pat} = {i_1, ..., i_N}$ represents the embeddings of $N$ image patches, $t_{cls}$ represents the embedding of the text \texttt{[CLS]} token, and $t_{tok} = {t_1, ..., t_M}$ represents the embeddings of $M$ text tokens.

\subsection{Manipulation Detection}
\label{ssec:MD}

Manipulation detection involves two tasks: fine-grained manipulation type detection and binary classification. The $\rm DGM^4$ dataset \cite{shao2023detecting} introduces two image manipulation methods: Face Swap (FS) and Face Attribute (FA), as well as two text manipulation methods: Text Swap (TS) and Text Attribute (TA). In HAMMER \cite{shao2023detecting}, the embedding of the \texttt{[CLS]} token for binary classification of the four manipulation types. However, this approach may suffer from modality competition \cite{huang2022modality}, where one modality is learned well while the other is not fully explored. For instance, text information may dominate \texttt{[CLS]} token, making it challenging to optimize the image part and leading to more imbalance modality information in \texttt{[CLS]} token. To address this issue and promote modal specificity, we introduce decoupled fine-grained classifiers (DFC) to independently guide visual and linguistic features. DFC consists of $C_i$ and $C_t$, where $C_i$ outputs one of the three categories of Real/FS/FA, and $C_t$ outputs one of the three categories of Real/TS/TA. $i_{cls}$ and $t_{cls}$ are separately fed into $C_i$ and $C_t$. The fine-grained classification loss is computed as follows:

\begin{equation}
\mathcal{L}_{mcls} = \mathcal{L}_{ce}(C_{i}(i_{cls}), y_{i}) + \mathcal{L}_{ce}(C{t}(t_{cls}), y_{t}),
\end{equation}
where $\mathcal{L}_{ce}$ denotes the cross-entropy loss, $y_i$ represents the image's fine-grained classification label, and $y_t$ represents the text's fine-grained classification label. Additionally, we concatenate $i_{cls}$ and $t_{cls}$ and feed them to a binary classifier $C_b$, computing the binary classification loss. Here, $y_b$ represents the binary classification label:

\begin{equation}
\mathcal{L}_{bcls} = \mathcal{L}_{ce}(C_{b}(\{i_{cls}, t_{cls}\}), y_{b}).
\end{equation}

\subsection{Manipulation Grounding}
\label{ssec:MG}

Manipulation grounding involves two tasks: manipulated image grounding and manipulated text grounding. In manipulated image grounding, the goal is to output the coordinates of the manipulated region in the image. In manipulated text grounding, the objective is to classify each token in the text and determine if it is manipulated. Drawing inspiration from the works of DETR \cite{carion2020end} and MaskFormer \cite{cheng2021per}, we propose an implicit manipulation query (IMQ) module, which consists of two components: I-IMQ and T-IMQ. The IMQ module utilizes learnable queries to adaptively aggregate intra-modal forged clues and emphasize modality-specific features. Taking the text in Fig .1 as an example, there is a significant inconsistency between "MP", "underpants", and "award". T-IMQ can efficiently model relationships between tokens by leveraging implicit forgery features learned by queries during training. The image manipulation features $f_{im}$ and text manipulation features $f_{tm}$ are aggregated by image manipulation queries $q_{im}$ and text manipulation queries $q_{tm}$, respectively:

\begin{equation}
\begin{aligned}
f_{im} &= \text{Attention}(q_{im}, i_{pat}, i_{pat}), \\
f_{tm} &= \text{Attention}(q_{tm}, t_{tok}, t_{tok}).
\end{aligned}
\end{equation}

The $f_{im}$ is inputted into a bbox detector $D_i$ to estimate the manipulated coordinates. The manipulated image grounding loss is calculated by combining the L1 loss and GIoU loss \cite{rezatofighi2019generalized}, where $y_{mig}$ represents the manipulated image grounding label:

\begin{equation}
\mathcal{L}_{mig} = \mathcal{L}_{L1}(D_i(f_i) - y_{mig}) + \mathcal{L}_{GIoU}(D_i(f_i) - y_{mig}).
\end{equation}

The $f_{tm}$ and $t_{tok}$ are dimensionally reduced, and an inner product is performed in the feature dimension to predict whether each token is manipulated. The manipulated text grounding loss is calculated using the cross-entropy loss, where $y_{mtg} = \{y_i\}^M_{i=1}$ and $y_i \in {0, 1}$ denotes whether the $i$-th token is manipulated or not:

\begin{equation}
\mathcal{L}_{mtg} = \mathcal{L}_{ce}(t_{tok} \times f_{tm}^T, y_{mtg}).
\end{equation}

\subsection{Loss function}
\label{ssec:Loss}

To obtain the final loss function, we combine the above components, where $\alpha$, $\beta$, and $\gamma$ are hyperparameters that control the relative importance of each loss term:

\begin{equation}
\mathcal{L} = \mathcal{L}_{bcls} + \alpha\mathcal{L}_{mcls} + \beta\mathcal{L}_{mig} + \gamma\mathcal{L}_{mtg}.
\end{equation}

\section{experiment}
\label{sec:experiment}

\subsection{Implementation details}
\label{ssec:detail}

The length of the text content is padded or truncated to 50 tokens, while the images are resized to 256x256 pixels. The uni-modal encoders are implemented by ViT-B/16 and RoBERTa. The modality interaction module is constructed using 6 transformer layers. The pre-training weights are derived from METER \cite{dou2022empirical}. The binary classifier, fine-grained classifier, and bbox detector are all implemented using multi-layer perceptron layers. The coefficients of the loss function are set as $\alpha = 1, \beta = 0.1, \gamma = 1$. We utilize the AdamW \cite{loshchilov2017decoupled} optimizer with a weight decay of 0.02. During the first 1000 steps, the learning rate is warmed up to 1e-4 and then decayed to 1e-6 using a cosine schedule.

\begin{table*}[t]
\centering
\caption{Results comparison with state-of-the-art methods and multi-modal learning methods on $\rm DGM^4$.}
\resizebox{0.9\textwidth}{!}{
\begin{tabular}{cc|ccc|ccc|ccc|ccc}
\multicolumn{2}{c|}{Categories} & \multicolumn{3}{c|}{Binary Cls}    & \multicolumn{3}{c|}{Multi-Label Cls}  & \multicolumn{3}{c|}{Image Grounding} & \multicolumn{3}{c}{Text Grounding} \\ \hline
Methods     & Params                    & AUC   & EER   & ACC   & mAP   & CF1   & OF1   & IoUmean & IoU50 & IoU75 & Precision & Recall & F1    \\ \hline
CLIP~\cite{radford2021learning} & - & 83.22 & 24.61 & 76.40 & 66.00 & 59.52 & 62.31 & 49.51   & 50.03 & 38.79 & 58.12     & 22.11  & 32.03 \\
ViLT~\cite{kim2021vilt}  & -       & 85.16 & 22.88 & 78.38 & 72.37 & 66.14 & 66.00 & 59.32   & 65.18 & 48.10 & 66.48     & 49.88  & 57.00 \\
HAMMER~\cite{shao2023detecting} & 441M & 93.19 & 14.10 & 86.39 & 86.22 & 79.37 & 80.37 & 76.45   & 83.75 & 76.06 & 75.01     & 68.02  & 71.35 \\
\textbf{Ours} & 328M & \textbf{95.11} & \textbf{11.36} & \textbf{88.75} & \textbf{91.42} & \textbf{83.60} & \textbf{84.38} & \textbf{80.83}  & \textbf{88.35} & \textbf{80.39} & \textbf{76.51} & \textbf{70.61} & \textbf{73.44} \\
\end{tabular}
}
\label{tab:sota}
\end{table*}

\begin{table}[htb]
\centering
\vspace*{-4mm}
\caption{Ablation study on DCA, DFC, and IMQ.}
\resizebox{0.45\textwidth}{!}{
\begin{tabular}{c|c|c|c|c|c} 
% Categories          & Binary Cls        & Multi-Label Cls   & Image Grounding   & Text Grounding    & Overall \\
% \hline
Methods             & AUC               & mAP               & IoUmean           & F1                & Avg     \\
\hline
Baseline            & 94.25             & 89.29             & 77.19             & 69.13             & 82.47   \\
\hline
Ours w/o $\rm M_t$  & 93.73             & 80.66             & \textbf{81.58}    & 46.58             & 75.64   \\
\hline
Ours w/o DFC        & 94.94             & 90.13             & 80.43             & \textbf{73.51}    & 84.75   \\
\hline
Ours w/o I-IMQ      & 95.05             & \underline{91.32} & 79.14             & 73.39             & 84.73   \\
Ours w/o T-IMQ      & \textbf{95.20}    & 90.66             & 80.75             & 73.09             & \underline{84.93} \\
\hline
\textbf{Ours}       & \underline{95.11} & \textbf{91.42}    & \underline{80.84} & \underline{73.44} & \textbf{85.20}    \\
\end{tabular}
}
\label{tab:ablation}
\end{table}

\subsection{Datasets and Evaluation Metrics}
\label{ssec:DS}

We conduct experiments on $\rm DGM^4$ \cite{shao2023detecting} dataset. The $\rm DGM^4$ comprise a total of 230k news samples, including 77,426 original image-text pairs and 152,574 manipulation pairs.

We evaluate each method using a total of twelve metrics across four tasks. For binary classification, we evaluate accuracy (ACC), area under the receiver operating characteristic curve (AUC), and equal error rate (EER). For fine-grained classification, we evaluate mean average precision (MAP), average per-class F1 (CF1), and overall F1 (OF1). For manipulated image grounding, we evaluate the mean intersection over union (IoUmean) as well as the IoU at thresholds of 0.5 (IoU50) and 0.75 (IoU75). For manipulated text grounding, we evaluate precision, recall, and F1 score.

\subsection{Comparison with the state-of-the-art methods}
\label{ssec:sota}

In this section, we show the performance of the state-of-the-art methods and our method on the $\rm DGM^4$ dataset. The comparison results are shown in Table 1. It can be observed that our method outperforms the state-of-the-art methods on the $\rm DGM^4$ dataset in all metrics. Particularly, compared to the SOTA method, our approach achieves improvements of over 2\% in important metrics such as ACC, MAP, IoUmean, and F1. These results indicate that our approach can effectively leverage modality-specific features and accurately model the correlations between images and text, leading to enhanced manipulation detection and grounding.

\subsection{Ablation Study}
\label{ssec:ablation}

To validate the importance of the DCA, DFC, and IMQ in our model, we conduct a series of ablation studies. We create the baseline model by removing DCA, DFC, and IMQ components. Specifically, we delete the $M_i$ branch, replace DFC with a multi-label classifier that makes predictions on concatenated features $\{i_{cls}, t_{cls}\}$, and remove IMQ while using MLP to locate the manipulation regions of image and text. By comparing the results to our full model, we can observe the effectiveness of our overall design. To verify the impact of each component, we perform ablation experiments by removing the corresponding component.

Removing $M_t$ (Ours w/o $\rm M_t$) leads to an average performance degradation. This indicates that the absence of DCA makes image features dominate and text features weakened. In addition, As mentioned in HAMMER \cite{shao2023detecting}, text manipulation detection and grounding are more difficult tasks than images. As a result, the relevant data decreases significantly. Furthermore, replacing DFC with a multi-label classifier (Ours w/o DFC) results in performance degradation on the classification task. This finding highlights the importance of DFC in enhancing modality-specific feature mining, enabling more accurate identification of specific types of manipulation. Moreover, the performance using IMQ outperforms its ablated counterparts (Ours w/o I-IMQ, Ours w/o T-IMQ) on the corresponding tasks. This demonstrates the effectiveness of the IMQ in aggregating forged features within each modality.

\begin{figure}[t]
\centering
\includegraphics[width=0.4\textwidth]{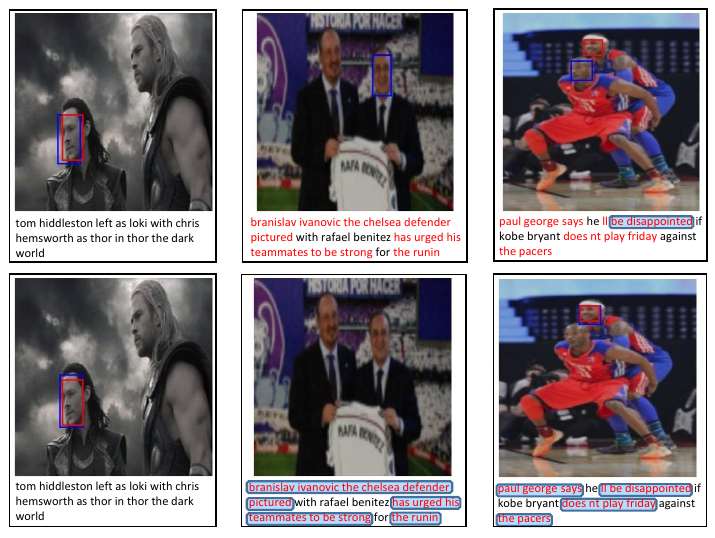}
\caption{Visualization of manipulation grounding results. Ground truths are in red, and predictions are in blue. The top three examples from HAMMER \cite{shao2023detecting}, and the subsequent three examples from our model.}
\label{Fig.vis}
\end{figure}

\subsection{Visualization}
\label{ssec:visual}

We provide visualizations of the manipulation grounding in Fig 2. In the second and third columns, we observe that HAMMER \cite{shao2023detecting} may encounter interference from the text modality, leading to errors in image grounding. In contrast, our method successfully distinguishes fake faces from real faces while accurately identifying the manipulated token. This illustrates the effectiveness of our approach in leveraging modality-specific features to uncover forged clues.

\section{CONCLUSION}
\label{sec:conclution}

In this paper, we construct a simple and novel transformer-based framework for manipulation detection and grounding. We introduce dual-branch cross-attention and decoupled fine-grained classifiers to effectively model cross-modal correlations and exploit modality-specific features. Implicit manipulation query is proposed to improve the discovery of forged clues. Experimental results on the $\rm DGM^4$ dataset show that our proposed approach outperforms existing methods in terms of performance.

\section{ACKNOWLEDGEMENT}
\label{sec:acknowledgement}

This work is supported by the National Natural Science Foundation of China (Grant No. 62121002).

% \vfill\pagebreak 

\bibliographystyle{IEEEbib}
\bibliography{strings,refs}

\section*{Supplementary Material}
\appendix

\section{Intra-domain and inter-domain comparison.} 

\begin{table}[htb]
\centering
\vspace*{-4mm}
\caption{Intra-domain and inter-domain comparison on $\rm DGM^4$ and Fakeddit.}
\resizebox{0.45\textwidth}{!}{
\begin{tabular}{c|c|c|c}
Datasets                        & $\rm DGM^4$  & Fakeddit & Overall \\ \hline
Methods                         & AUC   & AUC   & Avg   \\ \hline
HAMMER \cite{shao2023detecting} & 93.19 & 62.81 & 78.00 \\
Ours                            & \textbf{95.11} & \textbf{64.26} & \textbf{79.78} \\
\end{tabular}
}
\label{tab:generalization}
\end{table}

The field of multi-modal media manipulation detection and grounding currently only has the $\rm DGM^4$ dataset. To verify the effectiveness of our method on other datasets, we select the multi-modal fake news dataset Fakeddit with similar tasks and a large amount of data. We train the model on the DGM4 dataset and test it on the Fakeddit dataset. As shown in Table 3, the AUC of our model is higher than HAMMER \cite{shao2023detecting}, demonstrating the generalization ability of our method.

\end{document}